\documentclass[nofoot, twocolumn,10pt]{asme2e}

\title{Physics-Informed Neural Networks for Semiconductor Film Deposition: A Review}

\author{Tao Han\textsuperscript{1,2},  Zahra Taheri\textsuperscript{1},   Hyunwoong Ko\textsuperscript{1}\thanks{Corresponding Author.} \\   
    \affiliation{1. School of Manufacturing Systems and Networks\\
	Arizona State University\\
    2. Intel Corporation, AZ Factory\\
    }
}

\usepackage{amsmath}
\usepackage{amssymb}
\usepackage{graphicx}
\usepackage{epstopdf}
\usepackage{url}
\usepackage{hyperref}

\begin{document}

\maketitle  

\begin{abstract}
{\it 
Semiconductor manufacturing relies heavily on specialized film deposition processes, such as Chemical Vapor Deposition and Physical Vapor Deposition. These complex processes require precise control over critical parameters—including temperature, pressure, and material flow rates—to achieve film uniformity, proper adhesion, and desired functionality. Recent advancements in Physics-Informed Neural Networks (PINNs), an innovative machine learning (ML) approach, have shown significant promise in addressing challenges related to process control, quality assurance, and predictive modeling within semiconductor film deposition and other manufacturing domains. This paper provides a comprehensive review of ML applications specifically targeted at semiconductor film deposition processes. Through a thematic analysis, we identify key trends, existing limitations, and research gaps, offering insights into both the advantages and constraints of current methodologies. Reviewed studies are categorized into four key areas: (1) Process Control and Optimization; (2) Defect Image Recognition and Classification; (3) Tool Preventative Maintenance Prediction and Hardware Anomaly Detection; and (4) Atomic Layer Deposition Precursor Finding. Our structured analysis aims to highlight the potential integration of these ML techniques to enhance interpretability, accuracy, and robustness in film deposition processes. Additionally, we examine state-of-the-art PINN methods, discussing strategies for embedding physical knowledge, governing laws, and partial differential equations into advanced neural network architectures tailored for semiconductor manufacturing. Based on this detailed review, we propose novel research directions that integrate the strengths of PINNs to significantly advance film deposition processes. The contributions of this study include establishing a clear pathway for future research in integrating physics-informed ML frameworks, addressing existing methodological gaps, and ultimately improving precision, scalability, and operational efficiency within semiconductor manufacturing.}

\end{abstract}


\maketitle 

\vspace{-0.5em}
\section{Introduction}
\vspace{0.5em}
Semiconductor manufacturing relies on highly specialized processes for integrated circuit fabrication, with film deposition being a critical step \cite{Semiconprocessfundamentals}. Techniques such as Chemical Vapor Deposition (CVD) \cite{CVDSemiconprocessfundamentals} and Physical Vapor Deposition (PVD) \cite{PVDSemiconprocessfundamentals} are extensively used to create thin or thick films that function as conductive, insulating, or protective layers within semiconductor devices. These deposition processes demand precise control of various parameters, including temperature, pressure, and material flow rates, to ensure film uniformity, adhesion, and functionality \cite{HandbookofTFdepo}. The increasing demand for smaller, more efficient, and powerful semiconductor chips has intensified challenges in film deposition, necessitating advanced monitoring and control methods capable of managing complex interactions among process variables \cite{autocontrolinSMFG}.

Machine Learning (ML) \cite{whatisML} and Deep Learning (DL) \cite{whatisDL} have emerged as transformative tools to address these challenges. They provide capabilities to analyze extensive datasets, reveal underlying principles, enhance predictive accuracy, optimize process parameters, and enable real-time decision-making in semiconductor manufacturing \cite{ArtificialIntelligenceforDigitisingIndustryApplications}.
Recent advances in Physics-Informed Neural Networks (PINNs) \cite{physics-informedML, paper-PINNDLframework} represent particularly promising developments. PINNs integrate physics-based constraints directly into neural network architectures, ensuring model predictions align with established physical laws—an especially beneficial feature when experimental data are limited or expensive to obtain. 
These advanced neural network approaches hold significant potential for addressing challenges related to process control, quality assurance, and predictive modeling within semiconductor film deposition and other manufacturing applications.
    
To systematically address these issues, this paper conducts a comprehensive literature review of ML applications in film deposition processes, specifically within semiconductor manufacturing. Through thematic analysis, we identify key trends, limitations, and research gaps, providing insights into the strengths and constraints of existing methodologies. Reviewed studies are categorized into four major thematic areas: (1) Process Control and Optimization; (2) Defect Image Recognition and Classification; (3) Tool Preventative Maintenance Prediction and Hardware Anomaly Detection; and (4) Atomic Layer Deposition (ALD) Precursor Finding. By systematically analyzing the literature, we present a structured perspective on integrating these technologies to enhance interpretability, accuracy, and robustness in film deposition processes. Furthermore, we review state-of-the-art PINN principles and approaches, discussing methods for incorporating prior physics knowledge, physical laws, and partial differential equations (PDEs) into advanced neural networks for film deposition. Based on our analysis, we propose novel research directions combining the strengths of PINNs to significantly advance semiconductor manufacturing film deposition processes.

The remainder of this paper is structured as follows. Section 2 briefly introduces semiconductor manufacturing, film deposition processes, their challenges, and the potential of ML and DL technologies. Section 3 provides an in-depth literature review, emphasizing DL for film deposition and identifying existing limitations. Section 4 introduces PINNs and reviews relevant studies. Section 5 discusses future research directions, and Section 6 concludes the paper.
\vspace{-0.5em}
\section{Film Deposition in Semiconductor Manufacturing}
\vspace{0.5em}
\subsection{Overview of Semiconductor Manufacturing}
\vspace{0.5em}
Semiconductor manufacturing is a cornerstone of modern technology, encompassing the design, fabrication, and testing of microchips used in virtually all electronic devices \cite{introtosemiconMFg}. 
From smartphones and computers to medical devices and autonomous vehicles, semiconductors form the backbone of the digital era, enabling innovations that have transformed industries and everyday life. 
As transistor sizes approach atomic scales, several challenges threaten the sustainability of Moore’s Law \cite{Moorelaw}, for example, Quantum Effects, Materials and Design Limitations, Thermal Management, and Cost Increases. 
Despite challenges, the semiconductor industry continues to extend Moore’s Law through great innovations like Extreme Ultraviolet Lithography \cite{bakshi2009euv}, FinFET \cite{ding2022introFINFET} and Gate-All-Around Transistors \cite{GAAvsFINFET}, High-k/Metal Gate Materials\cite{frank2011HKMG}, 3D Integration and Chip Stacking, and Advanced Packaging\cite{lau2022Packagingreview}. 
With the rapid development of ML, it is expected to bring new breakthroughs and vitality to the continuation of Moore's Law.  

Semiconductor manufacturing is the process of creating Integrated Circuits (ICs) that form the core of modern electronic devices. 
These ICs are made from raw materials like silicon and contain billions of tiny transistors that control electrical signals. In current advanced technology manufacturing processes, it often operates at nanometer scales—dimensions far smaller than a human cell. Thus each step requires high quality control, extreme precision and cleanliness.

\vspace{-0.5em}
\subsection{Film Deposition Introduction and Challenges in Advanced Tech Nodes}\label{SubSec:FilmDepositionChallenge}
\vspace{0.5em}
Film deposition is one of the critical processes in semiconductor manufacturing, where thin or thick layers of materials—such as metals, dielectrics, or semiconductors-are deposited on a silicon wafer to meet the different functional requirements of integrated circuits. 
These films serve as conductive paths, insulating barriers, or structural layers, making them critical for the performance and reliability of semiconductor devices\cite{princofvapordepoTF}. 

Film deposition techniques can be broadly categorized into two primary categories: (1) PVD and (2) CVD (Fig. 1).
PVD is predominantly used for metal film deposition. 
This method physically transfers material from a solid target to the wafer surface through methods such as sputtering or evaporation under extreme vacuum conditions\cite{PVDofTF}. 
PVD forms conductive metal films that serve as interconnection layers in integrated circuits. 
CVD involves the reaction of precursor gases on the wafer’s surface under controlled temperature, pressure, and chemical conditions to form thin films \cite{CVDofTF}. 
Variants of CVD include 
(1) Plasma-Enhanced Chemical Vapor Deposition, which utilizes plasma to enable chemical reactions at lower temperatures, making it suitable for temperature-sensitive substrates, 
(2) ALD, a highly precise technique that deposits materials one atomic layer at a time, ensuring exceptional uniformity and thickness control, particularly on complex wafer surfaces \cite{ALDofTF}, and
(3) Epitaxy, a specialized CVD method that grows single-crystal layers aligning with the wafer's lattice structure, critical for high-performance transistor layers in advanced technology nodes.

\begin{figure*}[h]
    \centering
    \includegraphics[width=0.99\linewidth]{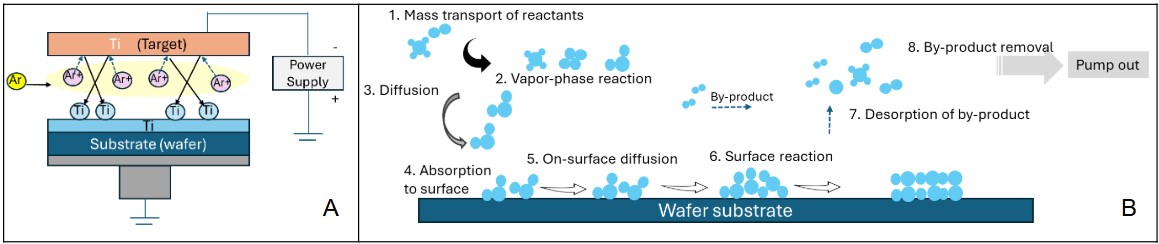}
    \caption{A. PVD; B. CVD. 
    In A, a Titanium (Ti) target serves as the source material and argon (Ar) gas is used to generate a plasma. The process occurs in a vacuum chamber, ensuring minimal contamination. The high-energy Ar ions bombard the Ti target, ejecting Ti atoms, which then deposit onto the substrate to form a Ti film\cite{PVDSemiconprocessfundamentals}. In B, precursor gases are introduced into the reaction chamber, where they undergo vapor-phase reactions. The reactive species then adsorb onto the wafer surface, facilitating surface reactions that lead to thin film formation. Subsequently, the resulting by-products are removed from the chamber through an exhaust system.\cite{CVDSemiconprocessfundamentals}. 
    }
    \label{fig:film depo}
\end{figure*}
\vspace{-0.5em}     
\subsection{Machine Learning and Deep Learning: Enabling Advanced Semiconductor Manufacturing}
\vspace{0.5em}
ML is a subset of AI that enables computers to identify patterns and make predictions based on data \cite{whatisML,trendsinDL}. 
Unlike traditional rule-based algorithms, ML models can autonomously learn complex relationships within datasets, making them ideal for process optimization and monitoring, such as defect detection, in semiconductor manufacturing.
DL is a specialized branch of ML that uses neural networks with multiple layers (hence ``deep") to model complex patterns and representations in large datasets. 
It has shown remarkable success in areas such as image recognition, process optimization, and predictive maintenance\cite{AIthroughIndustrialEquipment}.

Modern semiconductor manufacturing generates vast amounts of sensor, equipment, and production data,  
which ML and DL can leverage to uncover insights, predict outcomes, and improve decision-making as follows: 
(1) Process Optimization: ML algorithms can optimize manufacturing parameters to improve yield, reduce waste, and enhance efficiency. 
(2) Defect Detection: DL models, such as Convolutional Neural Networks (CNNs)\cite{CNN,paper2,paper5,paper7,paper8}, are adept at analyzing high-resolution images from inspection systems to detect defects at micro and nanoscale levels. 
(3) Predictive Maintenance: ML models analyze machine sensor data to predict equipment failures before they occur, minimizing downtime and reducing costs. 
(4) Precursor Finding: By synthesizing diverse data sources—including sensor readings, simulation outputs, and material properties—ML can provide a comprehensive view of manufacturing process dynamics to guide precursors finding.

As semiconductor manufacturing technology nodes become more complex and the critical dimension is scaling down to nanometer or even angstrom level (e.g., sub-5nm semiconductor devices), extremely tight process variation control is required. 
Thus more process tuning knobs are introduced, which make process control a more complex and challenging work.  Traditional methods struggle to model interactions and dependencies in film deposition processes, needing advanced machine learning deep approaches.
\vspace{-1em}
\section{Deep Learning for Film Deposition in Semiconductor Manufacturing}\label{DL_FilmDeposition}
\vspace{0.5em}
Neural network-based deep learning approaches, such as Bayesian optimization\cite{BRANN,paper1,Paper6,paper9,paper26}, reinforcement learning (RL)\cite{Reinforcement,paper9,paperz}, and CNN, have demonstrated significant potential for the challenges of film deposition introduced in Section \ref{SubSec:FilmDepositionChallenge}.
Based on a comprehensive review of the literature, this section provides a detailed examination of these applications, highlighting key studies, ML methodologies, results, and current limitations.
Existing DL studies in film deposition are categorized as (1) Process Control and Optimization, (2) Defect Image Recognition and Classification, (3) Tool Preventive Maintenance and Anomaly Detection, and (4)ALD Precursor Identification. 
\vspace{-0.5em}
\subsection{Process Control and Optimization} 
\vspace{0.5em}
Recent studies have increasingly leveraged advanced ML techniques to address complex optimization and modeling challenges in thin-film production processes. 
For example, artificial neural networks (ANNs) have been utilized effectively to estimate uncertain parameters within computationally demanding stochastic multiscale models for thin-film production via CVD \cite{paper0}.
Subsequent research has expanded the integration of ML with multiscale simulations. 
In particular, novel data-driven modeling frameworks combining Computational Fluid Dynamics (CFD) and ANNs have been developed to predict deposition rates accurately in ALD of silicon dioxide (SiO\textsubscript{2}) thin films. This approach achieves enhanced computational efficiency while maintaining high fidelity, demonstrating the practical advantages of hybrid modeling techniques \cite{paper1, paper16}.

Further advancements have incorporated optimization algorithms into ML frameworks. For instance, in optimizing the epitaxial growth of 4H–SiC via CVD processes, a combined approach utilizing Ant Colony Optimization (ACO) alongside Back Propagation Neural Networks (BPNNs) resulted in notable improvements, specifically a 17.2\% increase in deposition rate and a 51.88\% enhancement in uniformity compared to standard process conditions \cite{paper3}.
Complementing these methods, kinetic Monte Carlo (kMC) modeling guided by Density Functional Theory (DFT) calculations has been combined with Bayesian Regularized Artificial Neural Networks (BRANNs) to enhance modeling efficiency and accuracy in Plasma Enhanced Atomic Layer Deposition (PEALD) processes for HfO\textsubscript{2} thin films. This integrated modeling approach accurately simulates surface deposition mechanisms and efficiently predicts thin-film growth profiles, illustrating the benefit of coupling physics-based simulations with ML techniques \cite{Paper6}.

Recognizing dynamic variability in manufacturing environments, studies such as Run-Indexed Time-Varying Bayesian Optimization (RI-TVBO) have emerged to adaptively tune controllers for systems experiencing gradual shifts in process dynamics. This addresses a notable limitation of traditional auto-tuning methods, which often assume static system conditions over time \cite{paper9}.
Additionally, studies involving Vertical Gradient Freezing (VGF) crystal growth processes under varying gravitational conditions have demonstrated the potential of integrating Bayesian optimization with reinforcement learning for adaptive control strategies. These methods efficiently identify optimal process parameters and dynamically adjust control actions based on simulation feedback, showcasing the versatility and adaptability of ML in diverse manufacturing scenarios \cite{paper26}.

Collectively, these studies underscore the evolving synergy between ML and multiscale modeling, offering powerful tools to optimize and control complex film deposition processes with improved accuracy, efficiency, and adaptability.
\vspace{-1em}
\subsection{Defect Recognition and Classification}
\vspace{0.5em}
Recent studies have increasingly focused on leveraging DL techniques for defect recognition and classification in advanced manufacturing processes. For instance, the study by \cite{paper2} introduces DeepSEM-Net, a dual-branch DL architecture specifically developed for analyzing Scanning Electron Microscope (SEM) images of defects encountered in semiconductor manufacturing. By effectively addressing critical challenges related to defect segmentation and classification, DeepSEM-Net demonstrates significant improvements in accuracy compared to established models such as VGG16, ResNet34, and Swin-Transformer. This advancement highlights the importance of specialized architectures tailored to the intricacies of semiconductor defect analysis.

Expanding upon similar ML-based approaches, the work presented in \cite{paper5} utilizes deep belief networks (DBNs) to automatically characterize grain size and distribution in vanadium dioxide (VO\textsubscript{2}) thin films, based on atomic force microscope (AFM) images. This approach achieved an impressive 93.66\% accuracy in grain detection, illustrating the robust performance of ML techniques in accurately identifying microscopic material features critical to thin-film quality.

Addressing additional complexities inherent to defect recognition, study \cite{paper7} proposes the Deep Transfer Wasserstein Adversarial Network (DTWAN), a deep transfer learning model designed explicitly for wafer map defect recognition. DTWAN effectively mitigates common challenges in semiconductor manufacturing, such as limited labeled datasets and variable working conditions, by transferring knowledge learned from simulated wafer maps to real-world scenarios. Achieving a mean recognition rate of 86.50\% on target domain test data, DTWAN demonstrates superior performance compared to other transfer learning methods and conventional DL models. 

Taken together, these studies underscore the evolving capabilities of specialized deep learning frameworks in addressing diverse defect recognition challenges, substantially enhancing defect detection accuracy and reliability across film deposition processes.
 
\vspace{-0.5em}
\subsection{Tool Preventative Maintenance Prediction and Hardware Anomaly Detection}
\vspace{0.5em}
Recent research efforts have emphasized the use of ML methods for predictive maintenance and hardware anomaly detection, enhancing reliability and operational efficiency in manufacturing environments. For example, the study in \cite{paper4} investigates ML techniques specifically tailored for anomaly detection within Magnetron Sputtering (MS) processes, aiming to reduce failure rates and improve thin-film deposition efficiency. This research employs an offline anomaly detection strategy using Density-Based Spatial Clustering of Applications with Noise (DBSCAN) coupled with Dynamic Time Warping (DTW) to effectively identify abnormal patterns by comparing new deposition data against historical datasets from successful processes. Complementing this, an online anomaly detection approach leverages deep learning, specifically Long Short-Term Memory (LSTM) networks, to detect anomalies in real-time by identifying significant deviations between model predictions and observed process data.

Extending beyond hardware anomaly detection, the work presented in \cite{paper17} describes a contextual sensor system developed for non-intrusive machine status monitoring and energy consumption analysis. This system incorporates a Finite State Machine (FSM) model aligned with Standard Operating Procedures (SOP) to accurately capture interactions between workers and machines. Furthermore, it utilizes energy disaggregation methods to track the status and energy usage of individual machine components. 
\vspace{-0.5em}
\subsection{ALD Precursor Finding}
\vspace{0.5em}
Recent research has explored advanced data-driven methodologies for precursor identification in ALD. Study \cite{paper8} introduces a novel approach utilizing opinion mining from scientific literature through DL. The authors developed the Scientific Sentiment Network (SSNet), a DL model specifically designed to extract and classify expert opinions within scientific publications, with an emphasis on challenges and opportunities related to energy materials. This study diverges from previous citation-focused approaches by targeting the main body of scientific articles. SSNet integrates several sophisticated neural network architectures, including CNN, Attention-Based CNN (ABCNN), LSTM, and Collaborative Filtering (CF). The model achieves impressive performance metrics, reaching 94\% accuracy in opinion extraction (identifying opinion-bearing sentences) and 92\% accuracy in opinion classification (distinguishing between challenges and opportunities). Furthermore, SSNet's efficacy was rigorously compared against prominent large language models (LLMs), such as BERT, EnergyBERT, MatBERT, and GPT-4, with SSNet exhibiting comparable or superior performance.

A summary of the reviewed DL studies in film deposition is provided in Tables 1–4.

\begin{table*}[h]
    \caption{Process Control and Optimization}
    \label{tab:placeholder_label}
    \centering
    {
    \scriptsize
    \begin{tabular}{|p{1.0cm}|p{4.8cm}|p{3.0cm}|p{3.6cm}|p{3.8cm}|}
        \hline
        \rule{0pt}{2mm} 
        Study & Brief Analysis & ML Models & Results & Limitations \\ 
        \hline
        \rule{0pt}{6mm} 
        \cite{paper0} & ANN-based parameter estimation for stochastic multiscale thin-film CVD modeling. & Nonlinear autoregressive ANN & Successful real-time optimization and parameter estimation. & PDE model limitations; insufficient consideration of film uniformity on patterned surfaces. \\ 
        \hline
        \rule{0pt}{6mm}
        \cite{paper1, paper16} & Data-driven prediction of SiO$_2$ deposition rate in ALD processes. & Microscopic ANN, Macroscopic LPV Model. & Validation matches original CFD model performance. & Limited generalization; long-term accuracy concerns; simplified reaction kinetics. \\
        \hline
        \rule{0pt}{5mm}
        \cite{paper3} & Optimization of CVD process for 4H–SiC epitaxial growth. & ACO, BPNN & Good agreement between optimized parameters and experiments. & Simplified temperature and reaction models; narrow optimization scope.\\
        \hline
        \rule{0pt}{5mm}
        \cite{Paper6} & Modeling and optimization of PEALD process for HfO$_2$ films. & BRANN & Accurate reproduction of experimental growth rates and defects. & Simplified reaction mechanisms; limited structural and gas-phase considerations. \\
        \hline
        \rule{0pt}{5mm}
        \cite{paper9} & Automatic controller tuning via RI-TVBO for processes with gradual drift. & Bayesian Optimization, Gaussian Process & Effectively mitigates impacts of sudden drift changes. & No experimental validation; limited application scope. \\
        \hline
        \rule{0pt}{5mm}
        \cite{paper26} & VGF growth optimization for InGaSb crystals under varying gravity. & Bayesian Optimization, Reinforcement Learning & Effective diffusion coefficient estimation and gravity condition optimization. & Simplified 2D numerical model; limited experimental validation. \\
        \hline       
    \end{tabular}

    }
\end{table*}
\begin{table*}[h]
    \caption{Defect Recognition and Classification}
    \label{tab:placeholder_label}
    \centering
   {\scriptsize
    \begin{tabular}{|p{1.0cm}|p{4.8cm}|p{3.0cm}|p{3.6cm}|p{3.8cm}|}
        \hline
        \rule{0pt}{2mm}  Study & Brief Analysis & ML Models & Results & Limitations \\ 
        \hline
        \cite{paper2} & Introduces DeepSEM-Net, a dual-branch DL model for defect analysis using SEM images in semiconductor manufacturing. & DeepSEM-Net (outperforms VGG16, ResNet34, Swin-Transformer) & Achieves 97.25\% accuracy on a 5-class dataset. & Limited hyperparameter tuning; lacks extensive comparison with similar joint classification-segmentation models.\\
        \hline
        \cite{paper5} & Automates grain size and distribution characterization from AFM images in VO$_2$ thin films. & CNN, DBN, Deep Boltzmann Machines (DBM) & DBN achieves 93.66\% grain detection accuracy. & Challenges with complex grain shapes, limited model explainability, high computational cost on large datasets. \\
        \hline
        \cite{paper7} & Introduces DTWAN, a deep transfer learning model for wafer map defect recognition in semiconductor manufacturing. & DTWAN (ResNet, CNN, Wasserstein GAN) & Mean recognition rate of 86.50\% on target-domain test data; effective knowledge transfer from simulated to real data. & Limited real wafer data; computational complexity; potential overfitting; challenges with new and mixed defect types. \\
        \hline
    \end{tabular}
    }

\end{table*}

\begin{table*}[h]
   \caption{Tool Preventative Maintenance Prediction and ALD Precursor Finding}
    \label{tab:placeholder_label}
    \centering
     {\scriptsize
    \begin{tabular}{|p{1.0cm}|p{4.6cm}|p{3.0cm}|p{3.6cm}|p{4.0cm}|}
        \hline
        \rule{0pt}{2mm}  Study & Brief Analysis & ML Models & Results & Limitations \\ 
        \hline
        \cite{paper4} & Applies ML for anomaly detection in Magnetron Sputtering processes to enhance reliability. & DBSCAN, LSTM & Successful qualitative anomaly detection & Lacks quantitative evaluation metrics (accuracy, precision, recall). \\
        \hline
        \cite{paper17} & Develops a contextual sensor system for non-intrusive machine status and energy monitoring. & FSM-based SOP model & Effective implementation and evaluation on semiconductor fabrication equipment. & Limited detection capability, dependency on power signals, absence of experimental ML validation, limited scalability and single-machine focus. \\
        \hline
        \cite{paper8} & Proposes a novel knowledge-mining method using opinion-mining from scientific literature. & SSNet CNN, ABCNN, LSTM, CF & SSNet achieves 94\% accuracy in opinion extraction and 92\% in opinion classification. & Struggles with complex sentences containing multiple or contrasting opinions; limited to sentence-level sentiment analysis. \\
        \hline
    \end{tabular}
    }
 
\end{table*}   

\vspace{-0.5em}
\subsection{Challenges}
\vspace{0.5em}
Despite substantial efforts by numerous researchers and notable advances in applying ML and DL to semiconductor film deposition processes, critical challenges remain, hindering further development. 
One common limitation is the simplification of reaction mechanisms. 
For example, current Partial Differential Equation (PDE)-based models cannot clearly and comprehensively describe CVD reactions occurring on wafer surfaces. 
With advancements in semiconductor manufacturing technology, wafer surface conditions are becoming increasingly complex, characterized by smaller dimensions and the introduction of new materials. 
Consequently, existing simplified models struggle to accurately represent reactions under these evolving conditions.

Moreover, despite extensive data collection, many process parameters and their effects remain difficult to fully understand, interpret, and correlate with established outcomes. Although the results reported in reviewed studies are promising, they often lack clear explanations grounded in chemical and physical theories. 
Thus, it remains uncertain whether the applied conditions represent optimal choices. Collecting experimental data across a broader range of conditions would be both time-consuming and costly. 
Without clear theoretical guidance, the scope of experimental exploration is inherently limited.

\vspace{-1em}
\section{Physics-Informed Neural Networks}
\vspace{0.5em}
Physics-Informed Neural Networks (PINNs) train neural networks to comply with the mathematical representation of governing physics while utilizing datasets. 
PINNs integrate physics-based regularization terms, which constrain the solution space to align with physically plausible outcomes \cite{Azam2024PIGNN}. They address significant challenges in ML, such as generalization errors that violate physical principles when extrapolating beyond training data and the extensive reliance on labeled datasets \cite{Abbasi2024PAFs, Hao2022PIML}.
Compared to traditional methods, PINNs overcome the curse of dimensionality, seamlessly integrate experimental data with physical laws, generalize to new conditions, and eliminate the need for costly mesh generation. 
Their ability to handle both forward and inverse problems within a unified framework enhances efficiency and accuracy, particularly for solving complex and high-dimensional PDEs.



\vspace{-0.5em}
\subsection{Incorporating Physical Constraints in Neural Networks}
\vspace{0.5em} 
Physical constraints can be incorporated through strategies ranging from weak to strong forms influencing input data, model architecture, loss function, optimization, and inference mechanisms\cite{Hao2022PIML}. Among these techniques, emphasis is placed on observational data and physical loss functions, which offer accessible yet effective ways to integrate physical principles. Observational data follows a traditional approach, where machine learning models learn patterns directly from the data, indirectly capturing physical behaviors. Although this is a simpler method, it enables effective interpolation between data points influenced by physics. In contrast, physical loss functions provide a more advanced approach by embedding physical knowledge into the learning process. Additional terms in the loss function penalize violations of physical constraints, such as residuals of governing equations or conservation laws, ensuring predictions align with the underlying physical framework\cite{Abbasi2024PAFs}.
\vspace{-1.75em}
\subsubsection{Mathematical Framework for PINNs}
PINNs can be applied to various types of problems. In direct problems, the objective is to solve a differential equation subject to given initial and boundary conditions. A direct problem can generally be expressed as a set of equations, including the differential equation (e.g., equation \ref{eq:equation1}) and the initial or boundary conditions (e.g., equation \ref{eq:equation2}) as outlined in \cite{cuomo2022scientific}:
\setlength{\abovedisplayskip}{3pt}
\setlength{\belowdisplayskip}{3pt}
\setlength{\abovedisplayshortskip}{3pt}
\setlength{\belowdisplayshortskip}{3pt}
\begin{equation}
F(u(z);\gamma) = f(z), \quad z \in \Omega
\label{eq:equation1}
\end{equation}
\begin{equation}
B(u(z)) = g(z), \quad z \in \partial\Omega
\label{eq:equation2}
\end{equation}
where, \( z := [x_1, \dots, x_{d-1}, t] \) represents the spatiotemporal domain, \( \Omega \subset \mathbb{R}^d \) is the domain, \( \partial \Omega \) denotes the boundary, \( u \) is the unknown solution, \( \gamma \) represents the system’s physical parameters, \( f \) is the source term, \( F \) is the nonlinear differential operator, \( B \) indicates the initial or boundary conditions, and \( g \) is the boundary function. 

The loss function for equations \ref{eq:equation1} and \ref{eq:equation2} can be formulated as shown in equation \ref{eq:equation3}:
\begin{equation}
L = w_f L_f(\theta) + w_B L_B(\theta) + w_d L_d(\theta)
\label{eq:equation3}
\end{equation}
where \( \theta \) represents the model parameters. \( L_f \), \( L_B \), and \( L_d \) correspond to the PDE residual loss, boundary condition loss, and data loss, respectively. 
The weights \( w_f \), \( w_B \), and \( w_d \) are hyperparameters that balance the contributions of the different terms.

Another type of problem is the inverse problem. In these problems, unknown parameters or coefficients of the governing equations must be inferred from sparse or noisy observations. The loss function for inverse problems typically combines physical constraints with a data-based term that represents the mismatch between the predicted solution and the observed data\cite{raissi2020hidden, tartakovsky2020physics}. The solution process for this type of problem can be straightforward, as the two losses are consistent and can be minimized together without conflict \cite{lu2021physics}.

PDE-constrained inverse design is another class of problems, where the objective is to optimize a design while ensuring that the solution satisfies the governing PDE constraints, along with potential additional constraints such as manufacturing limitations. In general, this type of problem is formulated by a set of equations, given by equations \ref{eq:equation4}–\ref{eq:equation7}\cite{lu2021physics, Hao2022PIML}.
\begin{equation}
\min_{u, \zeta} J(u; \zeta)
\label{eq:equation4}
\end{equation}
Subject to
\begin{equation}
F[u(z); \zeta(z)] = 0, \quad z = (z_1, z_2, \dots, z_d) \in \Omega
\label{eq:equation5}
\end{equation}
\begin{equation}
B[u(z)] = 0, \quad z \in \partial\Omega
\label{eq:equation6}
\end{equation}
\begin{equation}
h(u, \zeta) = 0
\label{eq:equation7}
\end{equation}
where, \( J \) is the optimization function, \( F \) represents \( N \) different PDE operators \(\{F_1, F_2, \dots, F_N\}\), \( B \) denotes a general boundary condition operator, \( u(z) = (u_1 (z), u_2 (z), \dots, u_n (z)) \in \mathbb{R}^n \) is the solution of the PDEs, determined by \( \zeta(z) \), the quantity of interest. The last equation accounts for any additional constraints. As a result, the loss function, given by equation \ref{eq:equation8}, incorporates the optimization objective and the constraints outlined in equations \ref{eq:equation4}–\ref{eq:equation7}:
\begin{equation}
L = J + w_f L_f + w_B L_B + w_h L_h
\label{eq:equation8}
\end{equation}
where, \( L_f \), \( L_B \), and \( L_d \) represent the PDE residual loss, boundary condition loss, and data loss, respectively. The weights \( w_f \), \( w_B \), and \( w_d \) are used to balance the contributions of these terms.

Although equation \ref{eq:equation3} provides a versatile and effective framework, it often fails to approximate the exact latent solution accurately due to an imbalance in the gradients of the boundary condition loss term, initial condition loss term, and the PDE residual loss term during the training process. 
For example, the gradients of the boundary loss term may be significantly smaller in magnitude compared to the gradients of the PDE residual loss term. 
Consequently, the training process becomes dominated by minimizing the PDE residual loss, resulting in substantial inaccuracies in satisfying the boundary conditions.
Due to this imbalance, the model predicts solutions that adequately satisfy the PDE but struggle to meet the boundary requirements, ultimately resulting in inaccurate predictions \cite{wang2021understanding}. 
Such situations frequently occur for  PDE-constrained optimization problems and scenarios described by equation \ref{eq:equation8}, where the PDE-based loss (enforcing physics) and the objective function (guiding optimization) often conflict. 
As a result, the optimization process converges to solutions that typically do not satisfy the PDE constraints \cite{lu2021physics}. 

Determining loss weights is problem-specific and cannot rely on through trial-and-error as it is time-consuming and resource-intensive \cite{wang2021understanding}. 
To address this issue, two approaches are available: weighting strategies, where loss weights are adjusted to balance the contributions from different terms in the loss function, and adaptive resampling methods, which modify the training data distribution to emphasize regions exhibiting higher errors.
\subsubsection{Loss Reweighting Strategies}
\vspace{-0.5em}
Various methods, ranging from simple to sophisticated, have been developed to determine appropriate values for loss weights. Below, we highlight several prominent strategies.

Wang et al. \cite{wang2021understanding} propose an adaptive learning rate annealing method that automatically balances the interplay between different loss components by using gradient statistics . 
To balance the gradient magnitudes during back-propagation, the weights \( \lambda_i \) are adaptively updated using equations \ref{eq:equation9} and \ref{eq:equation10}:
\begin{equation}
\hat{\lambda}_i = \frac{\max_{\theta} |\nabla_{\theta} L_r (\theta)|}{|\nabla_{\theta} L_i (\theta)|}
\label{eq:equation9}
\end{equation}
\begin{equation}
\lambda_i \leftarrow (1-\alpha) \lambda_i + \alpha \hat{\lambda}_i
\label{eq:equation10}
\end{equation}
where, \( L_r \) denotes the PDE residual loss and \( L_i \) represents data-fitting losses, such as those from initial or boundary conditions. 
This approach improved the relative prediction error by more than one order of magnitude.

Maddu et al. \cite{maddu2023inverse} consider training a PINN as a multi-objective optimization problem . They introduce Inverse Dirichlet weighting as a strategy to mitigate the issue of vanishing task-specific gradients. Unlike gradient magnitude-based weighting, this method adjusts loss function weights based on gradient variance. This is achieved through an update rule, as given in equations \ref{eq:equation11} and \ref{eq:equation12}, where the weight for each objective (loss term) is proportional to the inverse of the standard deviation of its gradient.
\begin{equation}
\hat{\lambda}_K (\tau) = \frac{\max\limits_{t=1,\dots,K} \big( \textbf{std} \big( \nabla_{\theta^{(t)}} L_t (T) \big) \big)}{\textbf{std} \big( \nabla_{\theta^{(t)}} L_K (T) \big)}
\label{eq:equation11}
\end{equation}
\begin{equation}
\lambda_K (\tau+1) = \alpha \lambda_K (\tau) + (1-\alpha) \hat{\lambda}_K (\tau)
\label{eq:equation12}
\end{equation}
where, \(\textbf{std}\) denotes the standard deviation, \(\theta\) represents the model parameters, \(K\) refers to the \(K\)th objective, and \(L\) denotes the loss term. This approach effectively stabilizes the training process, particularly in problems with high-frequency components or high-order derivatives.

McClenny et al. \cite{mcclenny2023self} assign self-adaptive mask functions to the loss components associated with PDE residuals, boundary, and initial conditions. These mask functions are updated alongside the network parameters, using gradient descent for the network weights and gradient ascent for the self-adaptive weights. As the network encounters challenging regions, the self-adaptive weights increase, directing more focus and computational resources to those areas. Combined with Stochastic Gradient Descent (SGD), this method outperforms the fixed-weight model, reducing the norm error from 0.2079 to 0.0295, although it still lags behind the results reported in \cite{wang2021understanding}. In this study, the fixed-weight method uses constant coefficients (1.0 for PDE loss, 5.0 for initial conditions, and 50.0 for other conditions), chosen to match the average values of the self-adaptive weights at the end of training .

Liu and Wang \cite{liu2021dual} propose a new formulation for physics-constrained neural networks using a minimax architecture, where the training process involves solving a saddle point problem: minimizing over the network weights and maximizing over the loss weight parameters. Instead of the traditional gradient descent-ascent algorithm for saddle points, they introduce a saddle point search algorithm called the Dual-Dimer method, which improves computational efficiency for high-dimensional problems.

\subsubsection{Adaptive Data Resampling}
\vspace{-0.5em} 
As discussed in Section 4.1.1, one approach to addressing the challenges of imbalance learning is adaptively re-sampling collocation points. One such approach involves creating quasi-random points with a high level of uniformity as collocation points. Some notable sampling strategies include Sobol sequences \cite{sobol1967distribution}, Halton sequences \cite{halton1960efficiency}, Hammersley sampling \cite{hammersley1960monte}, and Latin hypercube sampling \cite{mckay2000comparison}, to name but a few.

Another approach is importance sampling, which dynamically adjusts the sampling strategy to prioritize regions with higher error levels. Previous research \cite{katharopoulos2018not, alain2015variance} has shown that convergence can be improved by selecting training samples with probabilities proportional to the gradient norm of the loss function. Since computing gradients for all points at each iteration is expensive, Nabian et al. \cite{nabian2021efficient} propose an efficient approach by using the loss function value as an approximation of loss gradient value, as shown in equation \ref{eq:equation13}:
\begin{equation}
q_j^{(i)} \approx \frac{J(\theta^{(i)}; x_j)}{\sum_{j=1}^N J(\theta^{(i)}; x_j)}, \quad \forall j \in \{1, \dots, N\}
\label{eq:equation13}
\end{equation}
Here, \( q_j^{(i)} \) represents the sampling probability distribution for the \( j \)-th collocation point at the \( i \)-th iteration, \( J \) denotes the loss function, and \( \theta \) is the model parameters. To further reduce computational cost, a piecewise constant approximation of the loss function is employed. In this approach, only a subset of points (seeds) is used to estimate loss values. The loss at each collocation point is approximated using the nearest seed’s loss, which enables a more efficient sampling process.

Katharopoulos et al. \cite{katharopoulos2017biased} introduce an importance sampling method that, similar to previous work, adjusts the sample probability distribution to emphasize more challenging samples with higher loss values. In each iteration, the model’s parameters, $\theta$, are updated using SGD to minimize the average loss,  as shown in equation \ref{eq:equation14}:  
\begin{equation}
\theta_{t+1} = \theta_t - \eta \alpha_i \nabla_{\theta_t} L(\psi(x_i; \theta_t), y_i)
\label{eq:equation14}
\end{equation}
In this equation, $\eta$ represents the learning rate, $L$ denotes the loss function, $i$ represents a discrete random variable sampled from a probability distribution $P$ with probabilities $p_i$, and $\alpha_i$ is a sample weight. In exact importance sampling, $p_i$ and $\alpha_i$ are determined by equation \ref{eq:equation15}:  
\begin{equation}
\alpha_i = \frac{1}{N p_i}, \quad p_i \propto L(\psi(x_i; \theta_t), y_i)
\label{eq:equation15}
\end{equation}
In equation \ref{eq:equation15}, $\psi$ denotes the deep learning model. This paper highlights that minimizing the average loss may not always yield the best model, particularly in the presence of rare samples. To address this, the sample weights are modified as in equation \ref{eq:equation16} to prioritize rare samples and minimize the maximum loss.  
\begin{equation}
\alpha_i = \frac{1}{N p_i^K}, \quad K \in (-\infty,1]
\label{eq:equation16}
\end{equation}
The value of $K$ determines the emphasis on high-loss samples, where smaller values shift the focus toward minimizing the maximum loss. When $K=1$, it corresponds to unbiased importance sampling. To reduce the computational cost of calculating the loss for each sample, the paper introduces a separate, lightweight neural network to predict the loss. This model is trained simultaneously with the main model, significantly reducing the computational complexity of the importance sampling method.

Tang et al. \cite{tang2023das} propose the Deep Adaptive Sampling (DAS) method to reduce statistical errors caused by discretizing the loss function using the Monte Carlo approximation, particularly when the error is localized. The DAS method enhances Monte Carlo estimations by employing adaptive sampling techniques, such as importance sampling, which prioritize regions with larger residuals to improve accuracy. To achieve this, it utilizes a deep generative model, KRnet, for probability density function approximation and efficient sample generation. The integration of KRnet enables effective variance reduction by approximating the distribution induced by the residuals.
\vspace{-0.5em}
\section{Future Direction}
\vspace{0.5em} 
PIML is expected to significantly enhance its applicability and effectiveness in the complex, data-intensive field of semiconductor film deposition processes. The unique characteristics of film deposition, governed by established physical and chemical laws, make it particularly suitable for PIML implementation. By incorporating these foundational principles into ML models, PIML can more accurately represent underlying processes, even when working with limited or noisy datasets. Semiconductor film deposition generates vast amounts of data, but not all parameters are continuously monitored due to cost and technical constraints. PIML models can leverage physical insights to infer missing data or constrain predictions, thereby improving robustness in sparse data scenarios.

In CVD processes, thin films are deposited through chemical reactions. As reviewed, PIML models can integrate thermodynamic and kinetic principles to optimize deposition parameters (e.g., temperature, pressure, precursor concentration), achieving improved film uniformity and quality. Plasma-enhanced processes, such as PECVD and PEALD, are inherently complex due to intricate interactions among ions, electrons, and reactive species. PIML enables embedding fundamental plasma physics equations into ML frameworks, thereby enhancing predictions of deposition rates and uniformity.
In semiconductor manufacturing, PIML can utilize PINNs not only to model film deposition but also other processes like etching, to solve heat and mass transfer problems. Future research can adopt PINNs specifically within film deposition processes, combining detailed process and equipment knowledge to effectively manage large volumes of daily operational data. This integration of physical and chemical domain expertise can improve decision-making accuracy and operational efficiency throughout wafer fabrication. 

Another promising direction involves combining PINNs with advanced ML techniques such as Graph Neural Networks (GNNs)\cite{8253599} and Continuous Graph Neural Networks (CGNNs)\cite{GNN2}. 
This integrated approach utilizes multi-modal data fusion with embedded spatial-temporal physical constraints, deepening the foundational understanding of complex phenomena and improving process control precision. 
GNNs contribute by synthesizing diverse data sources to create highly representative simulations, enhancing model accuracy, especially where experimental data is limited. This generative capability also increases the scalability and adaptability of the framework, enabling effective management of process variations and unexpected conditions, thus enhancing prediction reliability. 
CGNNs introduce causal inference capabilities, further complementing PINNs' physics-based constraints, and delivering interpretable, resilient predictive models that elucidate causal relationships within manufacturing processes. 
This comprehensive approach optimizes deposition outcomes, increases yield rates, and reduces costs.

As semiconductor technology progresses toward feature sizes at nanometer or even angstrom scales, process complexity and associated challenges continue to increase. 
Semiconductor manufacturing currently relies heavily on experienced personnel for systematic data collection, processing, and evaluation, essential for efficiently addressing diverse routine operational issues. 
Notably, existing research often overlooks the potential of ML as a holistic decision-making tool for integrated systems within semiconductor manufacturing facilities. 
Current studies primarily neglect ML's capability to synthesize diverse information sources to produce comprehensive decisions for routine factory operations. 
Future research directions should therefore emphasize ML as a system-level integrator and explore hybrid methodologies that merge domain expertise with advanced ML techniques. These innovations will significantly enhance the predictive accuracy and resilience of manufacturing systems, addressing routine operational challenges while shaping next-generation semiconductor manufacturing processes for enhanced scalability, energy efficiency, and product quality.

\vspace{-0.5em}
\section{Concluding Remarks}
\vspace{0.5em}
In this study, we provided a comprehensive summary of machine learning models applied to semiconductor manufacturing film deposition processes. We presented an in-depth categorization of machine learning and deep learning methods utilized across various studies, highlighting their methodologies, capabilities, and results within the context of film deposition. Additionally, we identified significant limitations and key challenges associated with these models, including reliance on extensive data, limited interpretability, and difficulty in generalizing beyond training datasets. Furthermore, we reviewed the unique capabilities of Physics-Informed Neural Networks to systematically incorporate physical principles, governing laws, and partial differential equations into neural networks. Based on these potentials, we proposed several promising future directions.
In future work, we will further investigate semiconductor data for advanced spatiotemporal modeling.

\section*{Acknowledgments}
The authors would like to acknowledge the support provided by Intel Corporation and the Fulton Fellowship from Arizona State University.
\bibliographystyle{asmems4}


\begin{thebibliography}{10}

\bibitem{Semiconprocessfundamentals}
El-Kareh, B., and Hutter, L.~N., 2012.
\newblock {\em Fundamentals of semiconductor processing technology}.
\newblock Springer Science \& Business Media.

\bibitem{CVDSemiconprocessfundamentals}
El-Kareh, B., and Hutter, L.~N., 2012, page 88, Chapter 3.1.
\newblock {\em Fundamentals of semiconductor processing technology}.
\newblock Springer Science \& Business Media.

\bibitem{PVDSemiconprocessfundamentals}
El-Kareh, B., and Hutter, L.~N., 2012, page 135, Chapter 3.3.
\newblock {\em Fundamentals of semiconductor processing technology}.
\newblock Springer Science \& Business Media.

\bibitem{HandbookofTFdepo}
Seshan, K., and Schepis, D., 2018.
\newblock {\em Handbook of thin film deposition}.
\newblock William Andrew.

\bibitem{autocontrolinSMFG}
Edgar, T.~F., Butler, S.~W., Campbell, W., Pfeiffer, C., Bode, C., Hwang, S.~B., Balakrishnan, K., and Hahn, J., 2000.
\newblock ``Automatic control in microelectronics manufacturing: Practices, challenges, and possibilities''.
\newblock {\em Automatica, \textbf{ 36}}(11), pp.~1567--1603.

\bibitem{whatisML}
Bell, J., 2022.
\newblock ``What is machine learning?''.
\newblock {\em Machine learning and the city: applications in architecture and urban design}, pp.~207--216.

\bibitem{whatisDL}
Kelleher, J.~D., 2019.
\newblock {\em Deep learning}.
\newblock MIT press.

\bibitem{ArtificialIntelligenceforDigitisingIndustryApplications}
Edited By Ovidiu~Vermesan, Chapter authors Cristina De~Luca, B. L. W. S. S. A.-B. G. P. A. R. F. P. M. C. R.~J., 2021,1st Edition, Chapter AI in Semiconductor Industry.
\newblock {\em Artificial Intelligence for Digitising Industry – Applications}.
\newblock Applications (1st ed.). River Publishers.

\bibitem{physics-informedML}
Karniadakis, G.E., K. I. L. L. e.~a., 2021.
\newblock ``Physics-informed machine learning''.
\newblock {\em Nature Reviews Physics, \textbf{ 3}}, pp.~422--440.

\bibitem{paper-PINNDLframework}
Raissi, M., Perdikaris, P., and Karniadakis, G., 2019.
\newblock ``Physics-informed neural networks: A deep learning framework for solving forward and inverse problems involving nonlinear partial differential equations''.
\newblock {\em Journal of Computational Physics, \textbf{ 378}}, pp.~686--707.

\bibitem{introtosemiconMFg}
Chao, T., 2000.
\newblock {\em Introduction to semiconductor manufacturing technology}.
\newblock Prentice Hall New Jersey.

\bibitem{Moorelaw}
Gordon~Moore, h.-l., 1965.
\newblock Moore's law.
\newblock Accessed: 2024-11-05.

\bibitem{bakshi2009euv}
Bakshi, V., 2009.
\newblock ``Euv lithography''.

\bibitem{ding2022introFINFET}
Ding, H., Yuan, L., and Yin, B., 2022.
\newblock ``Introduction to finfet: Formation process, strengths, and future exploration''.
\newblock In EMIE 2022; The 2nd International Conference on Electronic Materials and Information Engineering, VDE, pp.~1--7.

\bibitem{GAAvsFINFET}
Nagy, D., Indalecio, G., GarcíA-Loureiro, A.~J., Elmessary, M.~A., Kalna, K., and Seoane, N., 2018.
\newblock ``Finfet versus gate-all-around nanowire fet: Performance, scaling, and variability''.
\newblock {\em IEEE Journal of the Electron Devices Society, \textbf{ 6}}, pp.~332--340.

\bibitem{frank2011HKMG}
Frank, M.~M., 2011.
\newblock ``High-k/metal gate innovations enabling continued cmos scaling''.
\newblock In 2011 Proceedings of the European Solid-State Device Research Conference (ESSDERC), IEEE, pp.~25--33.

\bibitem{lau2022Packagingreview}
Lau, J.~H., 2022.
\newblock ``Recent advances and trends in advanced packaging''.
\newblock {\em IEEE Transactions on Components, Packaging and Manufacturing Technology, \textbf{ 12}}(2), pp.~228--252.

\bibitem{princofvapordepoTF}
{Sree Harsha}, K., 2006.
\newblock ``Chapter 12 - structure and properties of films''.
\newblock In {\em Principles of Vapor Deposition of Thin Films}, K.~{Sree Harsha}, ed. Elsevier, Oxford, pp.~961--1072.

\bibitem{PVDofTF}
Mahan, J.~E., 2000.
\newblock {\em Physical vapor deposition of thin films}.

\bibitem{CVDofTF}
Crowell, J.~E., 2003.
\newblock ``Chemical methods of thin film deposition: Chemical vapor deposition, atomic layer deposition, and related technologies''.
\newblock {\em Journal of Vacuum Science \& Technology A: Vacuum, Surfaces, and Films, \textbf{ 21}}(5), pp.~S88--S95.

\bibitem{ALDofTF}
Oke, J.~A., and Jen, T.-C., 2022.
\newblock ``Atomic layer deposition and other thin film deposition techniques: from principles to film properties''.
\newblock {\em Journal of Materials Research and Technology, \textbf{ 21}}, pp.~2481--2514.

\bibitem{trendsinDL}
França, R.~P., {Borges Monteiro}, A.~C., Arthur, R., and Iano, Y., 2021.
\newblock ``Chapter 3 - an overview of deep learning in big data, image, and signal processing in the modern digital age''.
\newblock In {\em Trends in Deep Learning Methodologies}, V.~Piuri, S.~Raj, A.~Genovese, and R.~Srivastava, eds., Hybrid Computational Intelligence for Pattern Analysis. Academic Press, pp.~63--87.

\bibitem{AIthroughIndustrialEquipment}
Elahi, M., Afolaranmi, S.~O., Lastra, J. L.~M., and Garcia, J. A.~P., 2023.
\newblock ``A comprehensive literature review of the applications of ai techniques through the lifecycle of industrial equipment''.
\newblock {\em Discover Artificial Intelligence}.

\bibitem{CNN}
Antonopoulos, I., Robu, V., Couraud, B., Kirli, D., Norbu, S., Kiprakis, A., Flynn, D., Elizondo-Gonzalez, S., and Wattam, S., 2020.
\newblock ``Artificial intelligence and machine learning approaches to energy demand-side response: A systematic review''.
\newblock {\em Renewable and Sustainable Energy Reviews, \textbf{ 130}}, p.~109899.

\bibitem{paper2}
Qiao, Y., Mei, Z., Luo, Y., and Chen, Y., 2024.
\newblock ``Deepsem-net: Enhancing sem defect analysis in semiconductor manufacturing with a dual-branch cnn-transformer architecture''.
\newblock {\em Computers \& Industrial Engineering, \textbf{ 193}}, p.~110301.

\bibitem{paper5}
Zerrouki, N., {Zouina Ait-Djafer}, A., Harrou, F., Lafane, S., Abdelli-Messaci, S., and Sun, Y., 2024.
\newblock ``Image-driven machine learning for automatic characterization of grain size and distribution in smart vanadium dioxide thin films''.
\newblock {\em Measurement, \textbf{ 233}}, p.~114791.

\bibitem{paper7}
Yu, J., Li, S., Shen, Z., Wang, S., Liu, C., and Li, Q., 2021.
\newblock ``Deep transfer wasserstein adversarial network for wafer map defect recognition''.
\newblock {\em Computers \& Industrial Engineering, \textbf{ 161}}, p.~107679.

\bibitem{paper8}
Xie, T., Wan, Y., Wang, H., Østrøm, I., Wang, S., He, M., Deng, R., Wu, X., Grazian, C., Kit, C., and Hoex, B., 2024.
\newblock ``Opinion mining by convolutional neural networks for maximizing discoverability of nanomaterials''.
\newblock {\em Journal of Chemical Information and Modeling, \textbf{ 64}}(7), pp.~2746--2759.
\newblock PMID: 37982753.

\bibitem{BRANN}
Burden, F., and Winkler, D., 2009.
\newblock {\em Bayesian Regularization of Neural Networks}.
\newblock Humana Press, Totowa, NJ, pp.~23--42.

\bibitem{paper1}
Ding, Y., Zhang, Y., Ren, Y.~M., Orkoulas, G., and Christofides, P.~D., 2019.
\newblock ``Machine learning-based modeling and operation for ald of sio2 thin-films using data from a multiscale cfd simulation''.
\newblock {\em Chemical Engineering Research and Design, \textbf{ 151}}, pp.~131--145.

\bibitem{Paper6}
Ding, Y., Zhang, Y., Orkoulas, G., and Christofides, P.~D., 2020.
\newblock ``Microscopic modeling and optimal operation of plasma enhanced atomic layer deposition''.
\newblock {\em Chemical Engineering Research and Design, \textbf{ 159}}, pp.~439--454.

\bibitem{paper9}
Cho, K., Shao, K., and Mesbah, A., 2024.
\newblock ``Run-indexed time-varying bayesian optimization with positional encoding for auto-tuning of controllers: Application to a plasma-assisted deposition process with run-to-run drifts''.
\newblock {\em Computers \& Chemical Engineering, \textbf{ 185}}, p.~108653.

\bibitem{paper26}
Rachid, G., 2023.
\newblock ``A numerical investigation and optimization by machine learning for the growth of ingasb crystals with a flatter interface by vertical gradient freezing method under microgravity and normal gravity conditions''.
\newblock {\em Osaka University}.

\bibitem{Reinforcement}
Arulkumaran, K., Deisenroth, M.~P., Brundage, M., and Bharath, A.~A., 2017.
\newblock ``Deep reinforcement learning: A brief survey''.
\newblock {\em IEEE Signal Processing Magazine, \textbf{ 34}}(6), pp.~26--38.

\bibitem{paperz}
Chen, Y.-L., Sacchi, S., Dey, B., Blanco, V., Halder, S., Leray, P., and De~Gendt, S., 2024.
\newblock ``Exploring machine learning for semiconductor process optimization: A systematic review''.
\newblock {\em IEEE Transactions on Artificial Intelligence}, pp.~1--21.

\bibitem{paper0}
Kimaev, G., and Ricardez-Sandoval, L.~A., 2020.
\newblock ``Artificial neural network discrimination for parameter estimation and optimal product design of thin films manufactured by chemical vapor deposition''.
\newblock {\em The Journal of Physical Chemistry C, \textbf{ 124}}(34), pp.~18615--18627.

\bibitem{paper16}
Cho, K., Shao, K., and Mesbah, A., 2024.
\newblock ``Run-indexed time-varying bayesian optimization with positional encoding for auto-tuning of controllers: Application to a plasma-assisted deposition process with run-to-run drifts''.
\newblock {\em Computers \& Chemical Engineering, \textbf{ 185}}, p.~108653.

\bibitem{paper3}
Tang, Z., Zhao, S., Li, J., Zuo, Y., Tian, J., Tang, H., Fan, J., and Zhang, G., 2024.
\newblock ``Optimizing the chemical vapor deposition process of 4h–sic epitaxial layer growth with machine-learning-assisted multiphysics simulations''.
\newblock {\em Case Studies in Thermal Engineering, \textbf{ 59}}, p.~104507.

\bibitem{paper4}
Delchevalerie, V., {de Moor}, N., Rassinfosse, L., Haye, E., Frenay, B., and Lucas, S., 2024.
\newblock ``When magnetron sputtering deposition meets machine learning: Application to process anomaly detection''.
\newblock {\em Surface and Coatings Technology, \textbf{ 477}}, p.~130301.

\bibitem{paper17}
Ren, Y., and Li, G.-P., 2022.
\newblock ``A contextual sensor system for non-intrusive machine status and energy monitoring''.
\newblock {\em Journal of Manufacturing Systems, \textbf{ 62}}, pp.~87--101.

\bibitem{Azam2024PIGNN}
Azam, F., et~al., 2024.
\newblock ``Accelerating power flow calculations in lv networks using physics-informed graph neural networks''.
\newblock In 2024 IEEE PES Innovative Smart Grid Technologies Europe (ISGT EUROPE), IEEE.

\bibitem{Abbasi2024PAFs}
Abbasi, J., and Andersen, P.~ ., 2024.
\newblock ``Physical activation functions (pafs): An approach for more efficient induction of physics into physics-informed neural networks (pinns)''.
\newblock {\em Neurocomputing, \textbf{ 608}}, p.~128352.

\bibitem{Hao2022PIML}
Hao, Z., et~al., 2022.
\newblock ``Physics-informed machine learning: A survey on problems, methods and applications''.
\newblock {\em arXiv preprint, \textbf{ arXiv:2211.08064}}.

\bibitem{cuomo2022scientific}
Cuomo, S., et~al., 2022.
\newblock ``Scientific machine learning through physics--informed neural networks: Where we are and what’s next''.
\newblock {\em Journal of Scientific Computing, \textbf{ 92}}(3), p.~88.

\bibitem{raissi2020hidden}
Raissi, M., Yazdani, A., and Karniadakis, G.~E., 2020.
\newblock ``Hidden fluid mechanics: Learning velocity and pressure fields from flow visualizations''.
\newblock {\em Science, \textbf{ 367}}(6481), pp.~1026--1030.

\bibitem{tartakovsky2020physics}
Tartakovsky, A.~M., Marrero, Y., Kecojevic, B.~T., Yudovich, I.~B., and Bauer, P.~D., 2020.
\newblock ``Physics‐informed deep neural networks for learning parameters and constitutive relationships in subsurface flow problems''.
\newblock {\em Water Resources Research, \textbf{ 56}}(5), p.~e2019WR026731.

\bibitem{lu2021physics}
Lu, L., Jin, P., Pang, G., Zhang, Y., and Karniadakis, G.~E., 2021.
\newblock ``Physics-informed neural networks with hard constraints for inverse design''.
\newblock {\em SIAM Journal on Scientific Computing, \textbf{ 43}}(6), pp.~B1105--B1132.

\bibitem{wang2021understanding}
Wang, S., Teng, Y., and Perdikaris, P., 2021.
\newblock ``Understanding and mitigating gradient flow pathologies in physics-informed neural networks''.
\newblock {\em SIAM Journal on Scientific Computing, \textbf{ 43}}(5), pp.~A3055--A3081.

\bibitem{maddu2023inverse}
Maddu, S., Sturm, D., Müller, C.~L., and Sbalzarini, I.~F., 2023.
\newblock ``Inverse dirichlet weighting enables reliable training of physics-informed neural networks''.
\newblock {\em arXiv preprint, \textbf{ arXiv:2301.00000}}.
\newblock Available at \url{https://arxiv.org/abs/2301.00000}.

\bibitem{mcclenny2023self}
McClenny, L.~D., and Braga-Neto, U.~M., 2023.
\newblock ``Self-adaptive physics-informed neural networks''.
\newblock {\em Journal of Computational Physics, \textbf{ 474}}, p.~111722.

\bibitem{liu2021dual}
Liu, D., and Wang, Y., 2021.
\newblock ``A dual-dimer method for training physics-constrained neural networks with minimax architecture''.
\newblock {\em Neural Networks, \textbf{ 136}}, pp.~112--125.

\bibitem{sobol1967distribution}
Sobol, I.~M., 1967.
\newblock ``The distribution of points in a cube and the approximate evaluation of integrals''.
\newblock {\em USSR Computational Mathematics and Mathematical Physics, \textbf{ 7}}, pp.~86--112.

\bibitem{halton1960efficiency}
Halton, J.~H., 1960.
\newblock ``On the efficiency of certain quasi-random sequences of points in evaluating multi-dimensional integrals''.
\newblock {\em Numerische Mathematik, \textbf{ 2}}, pp.~84--90.

\bibitem{hammersley1960monte}
Hammersley, J.~M., 1960.
\newblock ``Monte carlo methods for solving multivariable problems''.
\newblock {\em Annals of the New York Academy of Sciences, \textbf{ 86}}(3), pp.~844--874.

\bibitem{mckay2000comparison}
McKay, M.~D., Beckman, R.~J., and Conover, W.~J., 2000.
\newblock ``A comparison of three methods for selecting values of input variables in the analysis of output from a computer code''.
\newblock {\em Technometrics, \textbf{ 42}}(1), pp.~55--61.

\bibitem{katharopoulos2018not}
Katharopoulos, A., and Fleuret, F., 2018.
\newblock ``Not all samples are created equal: Deep learning with importance sampling''.
\newblock In International Conference on Machine Learning (ICML), PMLR, pp.~2525--2534.

\bibitem{alain2015variance}
Alain, G., et~al., 2015.
\newblock ``Variance reduction in sgd by distributed importance sampling''.
\newblock {\em arXiv preprint arXiv:1511.06481}.

\bibitem{nabian2021efficient}
Nabian, M.~A., Gladstone, R.~J., and Meidani, H., 2021.
\newblock ``Efficient training of physics‐informed neural networks via importance sampling''.
\newblock {\em Computer‐Aided Civil and Infrastructure Engineering, \textbf{ 36}}(8), pp.~962--977.

\bibitem{katharopoulos2017biased}
Katharopoulos, A., and Fleuret, F., 2017.
\newblock ``Biased importance sampling for deep neural network training''.
\newblock {\em arXiv preprint arXiv:1706.00043}.

\bibitem{tang2023das}
Tang, K., Wan, X., and Yang, C., 2023.
\newblock ``Das-pinns: A deep adaptive sampling method for solving high-dimensional partial differential equations''.
\newblock {\em Journal of Computational Physics, \textbf{ 476}}, p.~111868.

\bibitem{8253599}
Creswell, A., White, T., Dumoulin, V., Arulkumaran, K., Sengupta, B., and Bharath, A.~A., 2018.
\newblock ``Generative adversarial networks: An overview''.
\newblock {\em IEEE Signal Processing Magazine, \textbf{ 35}}(1), pp.~53--65.

\bibitem{GNN2}
Jiang, W., Liu, H., and Xiong, H., 2023.
\newblock ``When graph neural network meets causality: Opportunities, methodologies and an outlook''.
\newblock {\em arXiv preprint arXiv:2312.12477}.

\end{thebibliography}

\end{document}